\documentclass[letterpaper, 10 pt, journal, twoside]{IEEEtran}

\usepackage{graphicx}
\usepackage{amsmath,amssymb,bm,amsfonts}
\usepackage{threeparttable}

\usepackage{multirow}
\usepackage{xcolor}
\newcommand{\norm}[1]{\left\lVert#1\right\rVert}

\newcommand{\etal}{\textit{et al.\ }}
\newcommand\sbullet[1][.5]{\mathbin{\vcenter{\hbox{\scalebox{#1}{$\bullet$}}}}}
\usepackage{gensymb}
\usepackage[symbol]{footmisc}

\title{
Direct Visual-Inertial Ego-Motion Estimation \\via Iterated Extended Kalman Filter
}

\author{Shangkun Zhong and Pakpong Chirarattananon
\thanks{Manuscript received: September, 8, 2019; Revised December, 11, 2019; Accepted January, 2, 2020.}
\thanks{This paper was recommended for publication by Editor Eric Marchand upon evaluation of the Associate Editor and Reviewers' comments.
This work was supported by the Research Grants Council of the Hong Kong Special Administrative Region of China (grant number CityU-11215117).}
\thanks{The authors are with the Department of Biomedical Engineering, City University of Hong Kong, Hong Kong SAR, China (emails: shanzhong4-c@my.cityu.edu.hk, pakpong.c@cityu.edu.hk).}%
}

\begin{document}

\maketitle
\markboth{IEEE Robotics and Automation Letters. Preprint Version. Accepted January, 2020}
{ZHONG \MakeLowercase{\textit{et al.}}: Direct Visual-Inertial Ego-Motion Estimation via Iterated Extended Kalman Filter
} 

\begin{abstract}

This letter proposes a reactive navigation strategy for recovering the altitude, translational velocity and orientation of Micro Aerial Vehicles. The main contribution lies in the direct and tight fusion of Inertial Measurement Unit (IMU) measurements with monocular feedback under an assumption of a single planar scene. An Iterated Extended Kalman Filter (IEKF) scheme is employed. The state prediction makes use of IMU readings while the state update relies directly on photometric feedback as measurements. Unlike feature-based methods, the photometric difference for the innovation term renders an inherent and robust data association process in a single step. The proposed approach is validated using real-world datasets. The results show that the proposed method offers better robustness, accuracy, and efficiency than a feature-based approach. Further investigation suggests that the accuracy of the flight velocity estimates from the proposed approach is comparable to those of two state-of-the-art Visual Inertial Systems (VINS) while the proposed framework is $\approx15-30$ times faster thanks to the omission of reconstruction and mapping.

\end{abstract}

\section{Introduction}

\IEEEPARstart{E}{fficient} and robust motion estimation plays a vital role in the operation of autonomous aerial robots. In recent years, several Visual-Inertial Systems have emerged as a framework for simultaneously recovering camera's motion and 3D map points by complementing visual sensors with IMU measurements. The visual-inertial motion estimation is one of the most intensively researched areas thanks to its accuracy, scalability and low cost. VINS, either optimization-based \cite{okvis, qin2017vins,  mur2017visual} or EKF-based \cite{Mourikis2007A,Stephan2013Monocular,RVIO}, are capable of providing precise state estimation as environmental map points and camera poses are incrementally refined over a prolonged period. However, drawbacks of VINS exist. The refinement of a large number of poses and landmarks is of high computational complexity. While the sparse structure of the normal equation in the bundle adjustment and the incremental technique have been exploited to reduce the computational load \cite{Liu_2018_CVPR}, they are still unsuitable for real-time application on small robots with limited computational power. In addition, the robustness of VINS is influenced by the ability to continuously track features over a long period. This brings about some susceptibility to rapid motion, low-textured scenes, and varying light conditions. 
\begin{figure}[t] 
    \centering
    \includegraphics[scale=0.92]{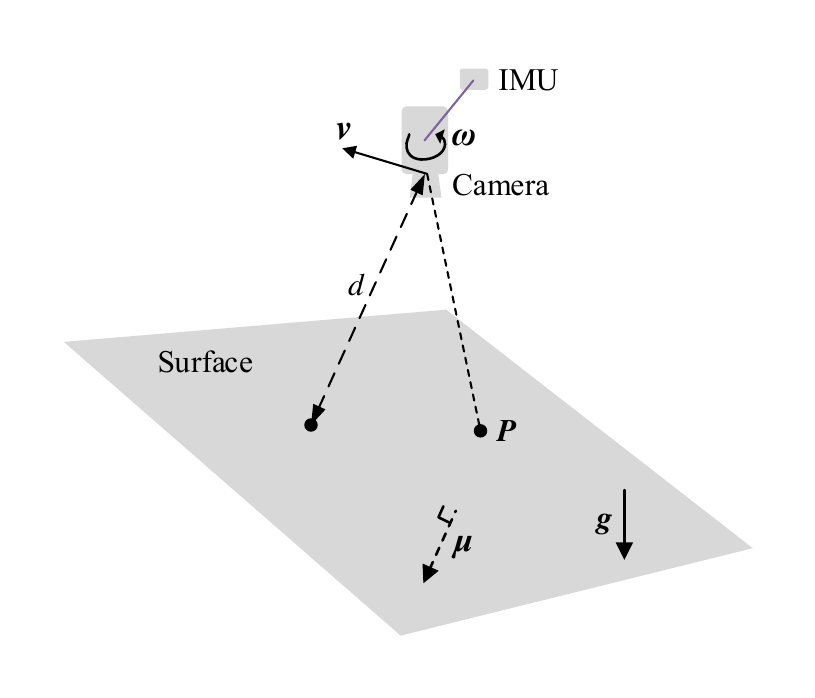}
    \caption{Diagram of an IMU-camera rig movement. The moving IMU-camera setup observes a single non-horizontal plane. The linear velocity $\bm{v}$, angular velocity $\bm{\omega}$, point $\bm{P}$ on the surface, the unit normal vector $\bm{\mu}$ and the normalized gravity vector $\bm{g}$ are expressed in the camera frame.}
    \label{fig:definitions}
\end{figure}

Another family of motion estimation methods is the reactive navigation. As a less demanding approach, reactive navigation only relies on the processing of most recent frames of images and sensory data. Most prevalently, it employs optical flow from the visual sensor to track features between consecutive frames \cite{Baker2004Lucas}. In contrast to the map-based navigation, the absence of landmarks' estimation significantly reduces the computational burden. Besides, optical flow-based methods offer more robust solutions as they are not required to maintain prolonged feature tracks. 

There have been several developments related to optical flow-based navigation. Izzo \etal presented a safe landing strategy using ventral optical flow and time-to-contact \cite{Izzo2012Landing}. Nevertheless, similar to VINS, uses of a monocular camera alone are unable to provide the metric scale. For this reason, Ho \etal proposed a distance and velocity estimator by taking into account the control inputs (instead of acceleration) to recover the metric scale \cite{DH_estimator_mono}. In another example, Grabe \etal incorporated onboard IMU data to recover the linear velocity and orthogonal distance to planar scenes by the formulation of a nonlinear observer under a single plane assumption  \cite{Grabe2015Nonlinear}. In the implementation, the plane's normal was assumed aligned with the gravity vector that is absolutely determined rather than estimated by the IMU measurements. To address the shortcoming, Hua \etal presented a nonlinear observer to estimate the depth, velocity, and gravity direction using the horizontal plane assumption \cite{hua2018attitude}. The observer is unable to handle inclined planes.

The aforementioned map-based VINS and reactive navigation methods rely on feature extraction and association processes such as the well-established Lucas-Kanade (LK) tracker \cite{Baker2004Lucas} to provide visual information. This feature identification and tracking process is relatively computationally demanding and may account for up to 99\% of the total processing time in reactive navigation as found in \cite{TRO_Adaptive_Gain_Control}. Alternatively, the \textit{direct} or \textit{featureless} method, which eliminates the feature detection and tracking process, has been proposed to further reduce the computational complexity. Using image intensity, direct implementations also enhance the robustness against motion blur or in scenes with little texture. Nevertheless, for map-based methods, intensive computation is required for the generation of dense depth-map \cite{newcombe2011dtam}. This issue is further remedied by the integration of feature tracking with patches of photometric feedback as a semi-direct approach  \cite{jin2003semi, rovio}.

\begin{figure}[tbp]
	\centering
	       \includegraphics[scale=0.8]{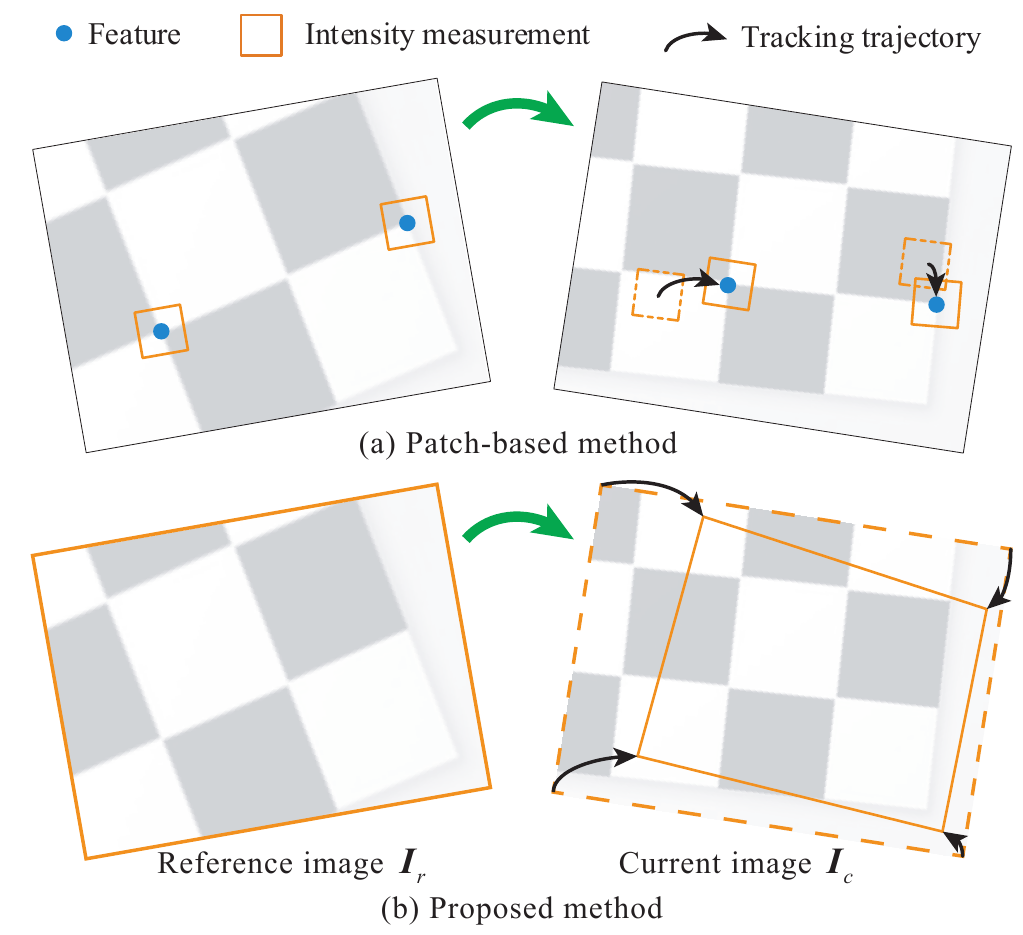}
    \vspace{-2mm}
	\caption{A sketch comparing the measurement model of the patch-based approach \cite{rovio} with the proposed method: (a) the patch-based method in \cite{rovio} and (b) the proposed method. In \cite{rovio}, an intensity patch around each image feature in the reference frame $\bm{I}_r$ is warped into the current frame $\bm{I}_c$ according to the feature's depth and relative pose between two frames provided by the IMU prediction. The difference between the warped patch and the actual measurement constitutes the innovation term to update the state vector. The black arrows denote the trajectory during the iterated update. In contrast, the proposed method aligns the entire reference image $\bm{I}_r$ to the current image $\bm{I}_c$ using the homography model under the single plane assumption. The image deformation is predicted by the camera's motion through the IMU integration.
	} \vspace{-2mm}
	\label{fig:pipeline}
\end{figure}

This paper presents a novel direct ego-motion estimation method for reactive navigation. Unlike previous direct methods for reactive navigation \cite{Zhang2016Bio, chirarattananon2018direct}, the proposed Iterated Extended Kalman Filter (IEKF) scheme efficiently estimates the inverse altitude, flight velocity, gravity direction, and plane's normal from photometric feedback in a single step. To achieve this, the single-plane assumption is employed. This is an attractive compromise when landmarks and the corresponding depth map is not considered. The assumption, also present in \cite{Grabe2015Nonlinear,hua2018attitude,Zhang2016Bio,chirarattananon2018direct}, radically simplifies the computation, eliminating the preference to consider image patches to reduce the computational complexity as found in recent map-based strategies \cite{jin2003semi,rovio}. Furthermore, the proposed approach has no restriction on the motion pattern of the camera or the plane's inclination as present in \cite{hua2018attitude,Zhang2016Bio,chirarattananon2018direct}.

The proposed framework takes motivation from the previous indirect reactive navigation method \cite{Grabe2015Nonlinear} and the semi-direct use of photometric feedback through IEKF \cite{rovio}. That is, the photometric error from an entire image is directly integrated into the IMU measurements for the ego-motion estimation via the IEKF framework under the assumption of a single planar scene. As illustrated in Fig. \ref{fig:pipeline}, the proposed method differs from the work \cite{rovio} owing to the single-plane assumption. The simplification means there exists only a low-dimension state vector associated with the image measurement model and the continuous homography equation. This makes the complex data association and mapping process unnecessary. To enhance the robustness, each pixel value on the image is integrated into Kalman update step instead of the multiple patches around the points of interest as in \cite{rovio}. To deal with the large observation vector comprising of pixel intensities from the entire image, a Gauss-Newton Kalman gain \cite{Bell1993The} is used to substantially reduce the computation complexity. The direct implementation yields an implicit and robust tracking process. Meanwhile, the gravity direction and plane's normal are independently estimated. To do so, a compact parameterization of bearing vectors on manifolds is employed, avoiding the singularity. IMU biases are also estimated.

The downsides of the proposed method exist. Compared with the popular VINS, the proposed system is less versatile as it cannot be applied when multiple planes exist in the view. Without mapping, the 3D position is not formulated. Nevertheless, the efficiency, robustness, and precision of our system serve as a potential surrogate for computationally constrained platforms, such as small and insect-scale flying robots \cite{shu2019quadrotor,chen2019controlled}.

The rest of this paper is structured as follows. Section \ref{preliminary} provides background on the continuous homography constraint. Section \ref{method} presents the IEKF formulation for directly estimating the inverse altitude, ratio velocity, planar normal vector and gravity direction from photometric feedback. In Section \ref{experiments}, extensive flight experiments were performed to evaluate and benchmark the performance of the proposed method with respect to two state-of-the-art VINS \cite{qin2017vins,rovio}. Lastly, conclusion and future directions are provided. 

\section{Continuous Homography Constraint \\and Optical Flow}{\label{preliminary}}

In this section, we briefly recall the derivation of the continuous homography constraint. Throughout the manuscript, vectors and matrices are represented by bold letters. Vectors are expressed with respect to the camera's frame unless stated otherwise. 

As illustrated in Fig. \ref{fig:definitions}, suppose a point $\bm{P}\in \mathbb{R}^3$ associated with a flat surface stationary in the inertia frame is observed by a moving camera with the linear velocity ${\bm v} \in \mathbb{R}^3$ and angular velocity $\bm{\omega} \in \mathbb{R}^3$. The motion of point $\bm P$ resulting from the camera movement is
\begin{equation}{\label{d_X_1}}
\dot{\bm P} = -\lbrack \bm{\omega} \rbrack_\times\bm{P}-\bm{v},
\end{equation}
where $\lbrack \bm{\omega} \rbrack_\times \in \mathbb{R}^{3\times3}$ denotes the skew-symmetric matrix associated with $\bm{\omega}$. Let ${\bm \mu} \in \mathbb{S}^2$ denote a unit vector normal to the plane, not necessarily parallel to the gravity direction $\bm{g}$, and $d = \bm{\mu}^T\bm{P}$ denote the orthogonal distance from camera center to the plane as illustrated in Fig. \ref{fig:definitions}. Eq. \eqref{d_X_1} becomes 
\begin{equation}{\label{d_X_with_n}}
\begin{split}
\dot{\bm P} &=  -(\lbrack \bm{\omega} \rbrack_\times+\dfrac{1}{d}\bm{v}\bm{\mu}^T)\bm{P}.
\end{split}
\end{equation}
Let $\bm{p} = \begin{bmatrix}
u,v,1
\end{bmatrix}^T$ be the projection of point $\bm{P}$ on the image plane and $\lambda$ denote depth of the point $\bm{P} $ in the camera frame ($\lambda > 0$), this yields
\begin{align}
\bm{P} &= \lambda \bm{M}^{-1}\bm{p}, \quad
\dot{\bm{P}} = \dot{\lambda}\bm{M}^{-1}\bm{p} + \lambda \bm{M}^{-1}\dot{\bm{p}},{\label{d_X_lambda}}
\end{align}
where $\bm{M}\in \mathbb{R}^{3\times3}$ is the pinhole camera intrinsic matrix. 
Substituting Eq. \eqref{d_X_lambda} into \eqref{d_X_with_n} provides
\begin{equation}
\begin{split}
\dot{\bm{p}} & = -\bm{H}\bm{p} - \frac{\dot{{\lambda}}}{{\lambda}}\bm{p},{\label{of_2_v_h}}
\end{split}
\end{equation}
where $ \bm{H} \in \mathbb{R}^{3\times 3} $ is known as the continuous homography matrix relating the optical flow $\dot{\bm{p}}$ to its coordinates $\bm{p}$ \cite{ma2012invitation}:
\begin{align}
    \bm{H} = \bm{M}(\lbrack \bm{\omega} \rbrack_\times+\dfrac{1}{d}\bm{v}\bm{\mu}^T)\bm{M}^{-1}.{\label{con_H}}
\end{align}
In Eq. \eqref{of_2_v_h}, since the third element of $\dot{\bm{p}}$ is always zero, we obtain
\begin{align}
\dot{{\lambda}}/{\lambda} &= -\bm{e}_z^T\bm{H}\bm{p},\label{dot_lambda_over_lambda}
\end{align}
where $\bm{e}_z = \begin{bmatrix} 0, 0, 1 \end{bmatrix}^T$. Substituting the result into Eq. \eqref{of_2_v_h} yields
\begin{align}
    \dot{\bm{p}} = -(\bm{1} - \bm{p} \bm{e}_z^T)\bm{H}\bm{p},{\label{optic_2_motion_plane}}
\end{align}
where $\bm{1}$ is a $3\times3$ identity matrix. Eq. \eqref{optic_2_motion_plane} relates the camera motion and orientation with respect to the plane to the optical flow $\dot{\bm{p}}$.

\section{IEKF Estimation Framework}{\label{method}}

In this section, we present the IEKF framework for estimating i) the distance to a flat plane, ii) the camera's translational velocity, iii) the plane's normal vector, and iv) the gravity direction. To achieve this, the IMU measurements are employed for propagation of the state and covariance estimates, while the photometric feedback from the camera is directly used for the correction of the predicted state. 
\subsection{State Definition}
The state vector consists of the following elements:
\begin{align}{\label{state_def}}
	\bm{x} := \left(
	\alpha, \bm{\vartheta}, \bm{\mu}_s, \bm{g}_s
	, \bm{b}_a, \bm{b}_\omega
	\right),
\end{align}
where $\alpha$ is the inverse orthogonal distance ($\alpha = d^{-1}$) from the camera center to the surface. The inverse parameterization has been shown to produce superior accuracy in \cite{rovio}. Similar to \cite{chirarattananon2018direct}, $\bm{\vartheta}:=\bm{v}/d \in \mathbb{R}^3$ is defined as the ratio of flight velocity to the distance. The unit normal vector $\bm{\mu}$ of the plane and the normalized gravity vector $\bm{g}$ are represented as members of manifolds on $\mathbb{S}^2$ \cite{rovio}. They can be obtained by rotating the basis vector $\bm{e}_z$ via rotations $\bm{\mu}_s, \bm{g}_s \in SO(3)$, such as $\bm{\mu} = \bm{\mu}_s(\bm{e}_z)$ and $\bm{g} = \bm{g}_s(\bm{e}_z)$. Compared to other parametrization method, such as azimuth and elevation, this implementation does not suffer from the singularity issue and it is relatively simple to derive their differentials. The separate treatment of $\bm{\mu}_s$ and $\bm{g}_s$ allows the estimation to deal with non-horizontal ground. The terms $\bm{b}_i$'s represent IMU biases as defined below. 

\subsection{State Prediction}\label{subsec:state_prop}
The state propagation begins with the discretization of the continuous dynamic model.
\subsubsection{State Dynamics}
The dynamics of the state is dependent on the specific acceleration $\hat{\bm{a}}$ and the angular rate $\hat{\bm{\omega}}$ of the camera frame. These quantities are related to the measurements from the accelerometer ($\bm{a}_m$) and gyroscope ($\bm{\omega}_m$). For simplicity, the IMU frame is assumed to be aligned with the camera's frame. In addition, the IMU readings are assumed to be corrupted by a bias $\bm{b}$ and white noise $\bm{w}$ such that
\begin{align}{\label{IMU_readings}}
	\bm{a}_m &= \hat{\bm{a}} + \bm{b}_a + \bm{w}_a, \quad 	\bm{\omega}_m = \hat{\bm{\omega}} + \bm{b}_\omega + \bm{w}_\omega.
\end{align}
Consequently, the state dynamics ($\dot{\bm{x}}$) are:
\begin{align}
\dot{\alpha} &= \alpha\bm{\mu}^T\bm{\vartheta}+w_{\alpha}, {\label{dalpha}} \\
\dot{\bm{\vartheta}} &= \alpha(\hat{\bm{a}} - g_0\bm{g})+(\bm{\mu}^T\bm{\bm{\vartheta}}\bm{1}-[\bm{\hat{\omega}}]_\times)\bm{\vartheta}+\bm{w}_{\vartheta}, {\label{d_vartheta}}\\ 
\dot{\bm{\mu}}_s &= \bm{N}(\bm{\mu}_s)^T\hat{\bm{\omega}}+\bm{w}_{\mu},\\
\dot{\bm{g}}_s &= \bm{N}(\bm{g}_s)^T\hat{\bm{\omega}}+\bm{w}_{g},\\
\dot{\bm{b}}_a &= \bm{w}_{b_a},\quad \dot{\bm{b}}_\omega = \bm{w}_{b_\omega},{\label{db_w}}
\end{align}
where $g_0 = 9.8$ ms$^{-2}$ is the gravitational acceleration. The terms $\bm{w}_i$'s are zero-mean Gaussian white noise. The operator $\bm{N}\left(\sbullet \right)$ linearly projects a $3\times1$ unit vector into the tangent space of a unit vector in $\mathbb{R}^2$ such that
\begin{align}
    \bm{N}(\bm{\mu}_s)=\begin{bmatrix}
    \bm{\mu}_s(\bm{e}_x), \bm{\mu}_s(\bm{e}_y)
    \end{bmatrix},
\end{align}
where $\bm{e}_x = \begin{bmatrix} 1, 0, 0 \end{bmatrix}^T$ and $\bm{e}_y = \begin{bmatrix} 0, 1, 0 \end{bmatrix}^T$ such that $\bm{\mu}_s(\bm{e}_{i})$'s are the basis vectors of the coordinate system. 
\subsubsection{Discretization}
The dynamics of the state described by the Eq. \eqref{dalpha}-\eqref{db_w} are nonlinear in nature. To leverage the IKEF, they are discretized using the forward Euler method:
\begin{align}
	\bm{x}_k^- \approx \bm{x}_{k-1}^+\boxplus\Delta T\dot{\bm{x}}_{k-1}^+.{\label{state_predict}}
\end{align}
where $k$ denotes the time index at instant $t_{k}$ and $\Delta T$ denotes the IMU sample time. $\bm{x}_{k-1}^{+}$ is an a-posteriori estimate at time $t_{k-1}$ and $\bm{x}_{k}^{-}$ an a-priori estimate at time $t_k$. The boxplus (or boxminus) operator in Eq. \eqref{state_predict} behaves as a regular addition (or subtraction) in the Euclidean space. The exception is when it is applied to unit vectors defined on 2-manifolds ($\mathbb{S}^2$). Readers are referred to \cite{rovio} for the detailed  definition of these operators. 

The propagation of the covariance matrix of the state uncertainty follows
\begin{align}
\bm{\Sigma}_{k}^{-}&=\bm{F}_{k-1}\bm{\Sigma}_{k-1}^{+}\bm{F}_{k-1}^T+\bm{G}_{k-1}\bm{Q}_{k-1}\bm{G}_{k-1}^T,{\label{predict_covariance}}
\end{align}
with $\bm{\Sigma}_{k-1}^{+}$ denoting an a-posteriori covariance at time $t_{k-1}$ and $\bm{\Sigma}_{k}^{-}$  an a-priori covariance at time $t_{k}$. $\bm{F}_{k-1}$ and $\bm{G}_{k-1}$ are the Jacobians of the propagated states with respect to the previous state $\bm{x}_{k-1}^{-}$ and process noise $\bm{w}_{k-1}$. $\bm{Q}_{k-1}$ is the covairance matrix of the additive process noise $\bm{w}_i$'s at $t_{k-1}$.

The state prediction is performed according to Eq. \eqref{state_predict} and \eqref{predict_covariance} at the rate determined by the frequency of the IMU measurements ($\Delta T^{-1}$), independent of the observation or visual feedback.\vspace{-2mm}

\subsection{Photometric Measurements}{\label{update}}
We directly use pixel intensities from the whole image as measurements for the state update. This constitutes one key contribution of our work.
\subsubsection{Image-based Measurement Model}
At time $t_{r}$, a point on the surface projected onto the reference image plane at $\bm{p}_{r}$ has the corresponding pixel intensities $\bm{I}_{r}(\bm{p}_{r})$ with $\bm{I}\in \mathbb{R}^{m\times n}$ denoting the 2D image domain. After a time period ($\delta T$), each spot displaces to a new location on the current image plane $\bm{I}_{c}$ according to the current state ($\bm{x}_k$). This motion can be described using a homography projective transformation mapping $\mathcal{H}_k(\bm{p}_{r}|\bm{x}_k)$. The corresponding pixel intensity remains identical under the constant brightness assumption:
\begin{align}{\label{ob_model}}
 \bm{I}_{r}(\bm{p}_{r}) = \bm{I}_{c}(\mathcal{H}_k(\bm{p}_{r}|\bm{x}_k)),
\end{align} 
where the mapping $\mathcal{H}_k(\bm{p}_{r}|\bm{x}_k)$ is derived from Eq. \eqref{optic_2_motion_plane},
\begin{equation}
\begin{split}
\mathcal{H}_k(\bm{p}_{r}|\bm{x}_k) \approx   \bm{p}_{r}-\delta T(\bm{1} - \bm{p}_{r}\bm{e}_z^T)\bm{H}_k(\bm{x}_k)\bm{p}_{r}.\label{eq:dense_mea_2}
\end{split}
\end{equation}
From here, we define an observation vector $\bm{z}_k \in \mathbb{R}^{mn}$ obtained by stacking the elements of $\bm{I}_{c}(\mathcal{H}_k(\bm{p}_{r}|\bm{x}_k))$ from the entire image. Let $\bm{h}(\bm{x}_k)$ be an observation model derived from Eq. \eqref{ob_model} and \eqref{eq:dense_mea_2}.  Subsequently, the measurement of pixel intensities over the entire image is modeled as
\begin{equation}
    \bm{z}_k = \bm{h}(\bm{x}_k)+ \bm{\eta}_k, \label{eq:measurement_vector}
\end{equation}
where $\bm{\eta}_k$ is the observation noise assumed to be zero-mean Gaussian white noise with covariance $\bm{R}_k$. This means the entire image is used as feedback for the state $\bm{x}_k$.
\subsubsection{Iterated State Update}
The update step is executed once a new image is available. Instead of relying on identified image features \cite{qin2017vins} or using the sparse patch-based intensity measurements as in \cite{rovio}, pixel intensities from the entire image are used as a measurement vector as outlined by Eq. \eqref{eq:measurement_vector}. In addition, the use of IEKF reduces the susceptibility to the inaccuracy of the initial estimate of the standard EKF. 

The update step is designed to find a Kalman gain that provides an approximate maximum a-posteriori probability estimate. This is equivalent to finding the a-posteriori estimate that minimizes the cost function
\begin{align}
\operatorname*{arg\,min}_{\bm{x}_{k}^+} \norm{\bm{x}_{k}^+ \boxminus \bm{x}_{k}^-}_{{\bm{\Sigma}_{k}^-}^{-1}}^2+\norm{\bm{z}_k- \bm{h}(\bm{x}_{k}^+)}_{\bm{R}_k^{-1}}^2{\label{cost_fun}}.
\end{align}
Since the dynamic model and measurement model are nonlinear, Eq. \eqref{cost_fun} is solved recursively via the IEKF framework. That is, each update step contains several iterative steps (denoted by a subscript $j$). Starting from $j=0$, the a-posteriori estimate of the state at the $j^{\text{th}}$ iteration is
\begin{align}
    \bm{x}_{k,j+1}^+ =& \bm{x}_{k,j}^{+}\boxplus \Delta \bm{x}_{k, j}.{\label{update_state}} \\
\Delta \bm{x}_{k, j} =& \bm{K}_{k, j}\left(\left (\bm{z}_k-\bm{h}(\bm{x}_{k,j}^+)\right ) +\bm{S}_{k,j}\bm{L}_{k,j}\left(\bm{x}_{k,j}^{+}\boxminus\bm{x}_{k}^{-}\right)\right)\nonumber\\
&-\bm{L}_{k,j}\left(\bm{x}_{k,j}^{+}\boxminus\bm{x}_{k}^{-}\right)\label{correction_term},
\end{align}
where matrices $\bm{L}_{k,j}$ and $\bm{S}_{k,j}$  are Jacobians \cite{rovio}:
\begin{align}
    \bm{L}_{k,j} = \dfrac{\partial{\bm{x}_k^-\boxplus\Delta\bm{x}}}{\partial \Delta\bm{x}}\left(\bm{x}_{k,j}^+\boxminus\bm{x}_k^-\right),\  \bm{S}_{k,j} = \dfrac{\partial{\bm{h}(\bm{x}_{k,j}^+)}}{\partial \bm{x}_{k,j}^{+}}.
\end{align}
Rather than using a regular Kalman gain for Eq. \eqref{correction_term}, we use the Gauss-Newton (GN) Kalman gain \cite{Bell1993The}:
\begin{align}
    \bm{K}_{k,j} = &
\left((\bm{L}_{k, j}^{T}\bm{\Sigma}_{k}^{-}\bm{L}_{k,j})^{-1}+\bm{S}_{k,j}^{T}\bm{R}_{k}^{-1}\bm{S}_{k,j}\right){}^{-1}\bm{S}_{k,j}^T\bm{R}_k^{-1}.{\label{kf_gain_gn}}
\end{align}
The use of GN gain dramatically improves the computational efficiency. The calculation of a standard Kalman gain is dominated by an inverse operation of an $mn\times mn$ matrix, which is overwhelmingly large for $m\times n$ image feedback. The inverse operation in Eq. \eqref{kf_gain_gn} is performed on a square matrix of which the dimension is determined by the length of $\bm{x}$ or $14\times14$. The complexity of Eq. \eqref{kf_gain_gn} is prevailed by the multiplication $\bm{S}_{k,j}$ or $O\{(mn)^2\}$ operations only.

Finally, the iteration is terminated when the 2-norm of $\Delta \bm{x}_{k,j}$ is below a certain threshold or the iteration reaches the maximum steps. The state covariance is updated only once with the Jacobians at the last $u^{\text{th}}$ iteration step according to
\begin{align}
    \bm{\Sigma}_{k}^+&=\bm{\Sigma}_{k}^- - \bm{K}_{k,u} \bm{S}_{k,u}\bm{L}_{k,u}^T\bm{\Sigma}_{k}^-\bm{L}_{k,u}.{\label{update_cov}}
\end{align}
Iterated updates effectively prevent the accumulation of errors and improve the convergence and accuracy, especially in the initialization with large uncertainty. On the other hand, the pre-defined termination conditions limit the iteration to a few steps, mitigating the extra computational burden.
\section{Experimental Evaluation}{\label{experiments}}
\begin{figure}[tbp]
	\centering
	\begin{tabular}{cc}
	       \includegraphics[scale=0.7]{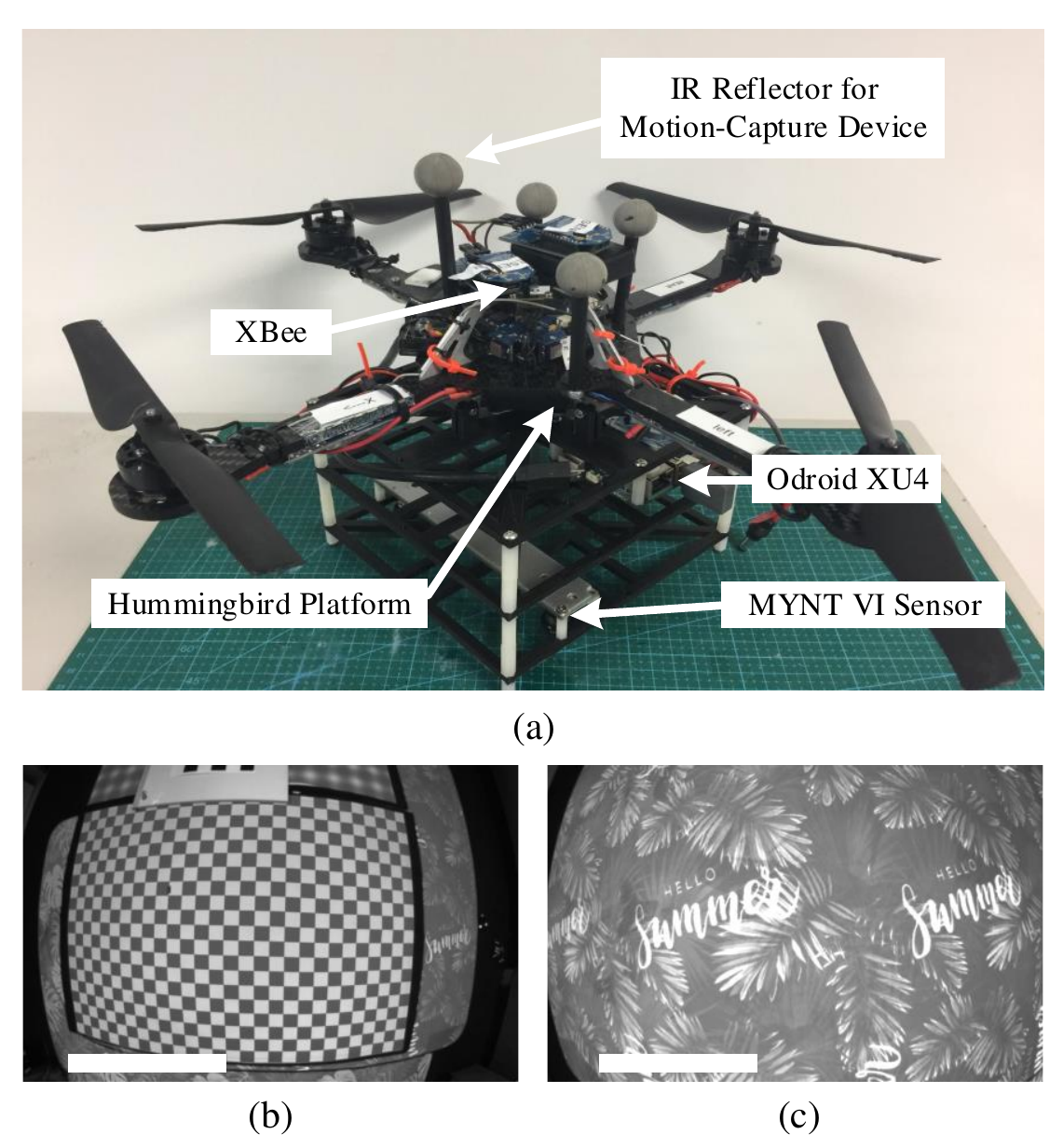}
	\end{tabular}
	
	\caption{(a) An AscTec Hummingbird quadrotor with a downward-facing MYNT VI sensor. (b), (c) two textures for experimental validation: Checkerboard (b) and Vegetation (c). The motion capture system is used for ground-truth measurements. Real-time control commands are transmitted from the ground station to the onboard controller via a pair of XBees. The white scale bars in (b) and (c) are 0.5 m and 0.3 m.}\vspace{-2mm}
	\label{fig:hummingbird}
\end{figure}
This section presents the results from several experiments to illustrate and assess our approach with various real-world datasets in terms of accuracy and computational cost. Root Mean Squared Errors (RMSE) of the estimated states with respect to the ground-truth are used for evaluation. In section \ref{compare_LK}, we first compare the proposed direct method with the traditional feature-based or LK method. Then a comparison between the proposed method and two state-of-the-art VINS is provided in section  \ref{ssec:compare_vins}. Lastly, the proposed method is tested when the robot flies over planes with different angles of inclination.

\subsection{Experimental Setup}

\newif\ifrmsv
\rmsvtrue
\begin{table*}[tbp]
	\centering
    \begin{threeparttable}
	\caption{Comparison of the estimation results from the D-IEKF, D-EKF and LK methods.}{\label{tab:results_lk_direct}}
	\begin{tabular}{c|c|c|c|c|c|c|c|c|c|c|c|c}
		\hline
		\multirow{2}{*}{Flight} & \multirow{2}{*}{Speed} & \multirow{2}{*}{Pattern} & {$\norm{\bm{v}}_{a}$,$\norm{\bm{v}}_{m}$}\tnote{1}  & \multicolumn{3}{c|}{Altitude RMSE (cm)} & \multicolumn{3}{c|}{Velocity RMSE (cm/s)} & \multicolumn{3}{c}{Average time cost (ms)} \\
	\cline{5-13}& &  & (cm/s)
	& D-IEKF & D-EKF & LK & D-IEKF & D-EKF & LK & D-IEKF & D-EKF & LK \\
		\hline
		\hline
	\textcircled{\footnotesize{1}}&  \multirow{5}{*}{Low}&CKB& \ifrmsv 24\else 39\fi,83 & \textbf{2.2} & {2.3} & 4.9 & 2.0 & 2.0 & \textbf{1.4}&{1.36}&\textbf{1.33} &6.97 \\
	\textcircled{\footnotesize{2}} &  & CKB & \ifrmsv21\else34\fi,113 & \textbf{3.7} & \textbf{3.7} & 7.3 &\textbf{1.2}&\textbf{1.2} & 1.8 & 1.35 & \textbf{1.34} & 7.01\\
	
	 \textcircled{\footnotesize{3}}&  & CKB & \ifrmsv 24\else 39\fi,94& \textbf{3.2} & \textbf{3.2}  & 73.6 & \textbf{2.2} & \textbf{2.2} & 118.5&1.37&\textbf{1.34}  &  6.91\\
	\textcircled{\footnotesize{4}} & & VEG & \ifrmsv21\else35\fi,64 & \textbf{2.9} & 3.6 &  5.5 & \textbf{1.4} & \textbf{1.4} &  \textbf{1.4} &  1.38& \textbf{1.33} & 7.11 \\
	\textcircled{\footnotesize{5}} & & VEG & \ifrmsv22\else38\fi,86& \textbf{3.2} & 5.7 & 29.7&\textbf{2.0}&2.2  & 35.4 & 1.37 & \textbf{1.34} & 7.13\\
		\cline{2-2}
    \textcircled{\footnotesize{6}} & \multirow{2}{*}{High} & VEG & \ifrmsv44\else72\fi,194 & \textbf{5.9} & 6.1 & 85.1 & \textbf{6.1} & 6.3 & 145.0 & 1.43 &\textbf{1.31}&7.25\\
	\textcircled{\footnotesize{7}}& & VEG & \ifrmsv57\else93\fi,275& \textbf{5.8} & *\tnote{2}  & 75.3& \textbf{7.0} & * & 146.2 & 1.44 &\textbf{0.63}&7.22\\
	\hline
	\end{tabular}
	\begin{tablenotes}
    \item [1] $\norm{\bm{v}}_{a}$ and $\norm{\bm{v}}_{m}$ are the root mean squared velocity and maximum velocity magnitude computed from the motion capture system used to describe the flight characteristics. 
    \item [2] The * symbol denotes a divergent estimate.
    \vspace{-4mm}
    \end{tablenotes}
	\end{threeparttable}
\end{table*}
For the experiments, we collected real-world datasets with an IMU-camera setup (MYNT AI, MYNT EYE) mounted on an AscTec Hummingbird quadrotor (Ascending Technologies) as shown in Fig. \ref{fig:hummingbird}(a). A motion tracking system (NaturalPoint, OptiTrack) was used to provide the ground-truth position and orientation, allowing the true state to be evaluated. For a horizontal surface, the true distance is the robot's altitude. The true velocity in the body frame is computed from the  position and then transformed into the body frame.

The visual-inertial sensor contains an ICM 2060 IMU from InvenSense and a MT9V034 camera from ON Semiconductor, both of which operate under hardware synchronization. Both intrinsics and extrinsics between these two sensors were calibrated beforehand. The IMU provides the measurements of specific accelerations and angular rates at 500 Hz. To attenuate the disturbance from vibration, a low-pass filter was employed. The IMU data were then downsampled to 100 Hz for the state prediction ($\Delta T= 0.01$ s). Grayscale images of size $752\times480$px were acquired at 30 frames per second. Both IMU measurements and images were recorded on the Odroid XU4 board and post-processed offline on a laptop with the Intel Core i5-8250U CPU at 1.6GHz. The offline implementation allows several estimation strategies to be compared using the same datasets. To verify the proposed estimation strategy, the algorithm was implemented in C++\footnote[2]{Available at https://github.com/ris-lab/direct-vi-iekf/}. Consecutive image frames are taken as the reference $\bm{I}_{r}$ and  current image $\bm{I}_{c}$ ($\delta T^{-1} = 30$ Hz).  All estimates were obtained with the same set of parameters unless specified. Assuming the state and measurement noises are statistically uncorrelated from one another and time independent, $\bm{Q}_k$ and $\bm{R}_k$ become diagonal and constant. The maximum iteration steps during the update stage (Eq. \eqref{cost_fun}) was set to three, the termination threshold of 2-norm of iteration change $\Delta \bm{x}_{k,j}$ to 0.05, the initial inverse altitude $\alpha_0$ to 10.0 m$^{-1}$, ratio velocity $\bm{\vartheta}_0$  to 0.0 s$^{-1}$. The initial normal vector $\bm{\mu}_0$ was chosen as $\left[0.2, -0.1, 0.97\right]^T$ to make the task more challenging and the initial gravity direction was set to $\left[0, 0, 1\right]^T$.

\subsection{Flights over Horizontal Ground}
For validation, we initially performed seven flights over two patterns on the horizontal ground and recorded the measurements. These patterns, Checkerboard (CKB) and Vegetation (VEG), shown in Fig. \ref{fig:hummingbird}(b)-(c), were selected as they feature salient corners and edges. During each flight, the robot was remotely controlled to follow an arbitrary trajectory covering an approximate $0.8\times0.8\times0.8$m volume for over 60 s. Among seven flights, five are low-speed flights with the RMS velocities of $\approx 0.2$ ms$^{-1}$ and the other two are high-speed flights with the RMS velocities over $0.4$ ms$^{-1}$. These two flight regimes were tested to inspect the performance of different estimation methods in different real-world scenarios. Outdoor flights are excluded due to the difficulty in obtaining ground-truth measurements and the single plane assumption may not hold in a complex landscape.

\subsubsection{Comparison of the Proposed Direct Method with the LK Method}{\label{compare_LK}}

\begin{figure}[htbp]
	\centering
    \begin{tabular}{cc}
    		\includegraphics[scale=0.4]{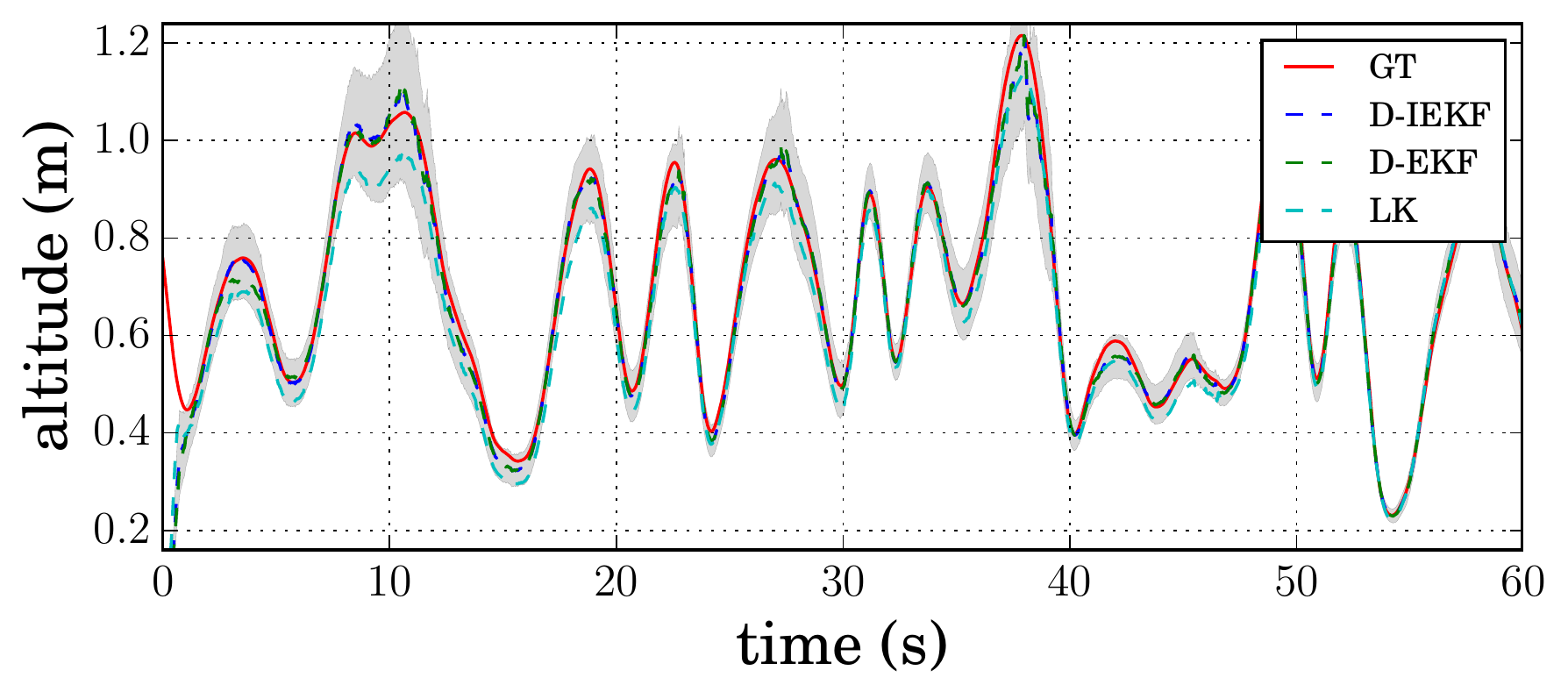} \\
    		(a)\\
    		\includegraphics[scale=0.4]{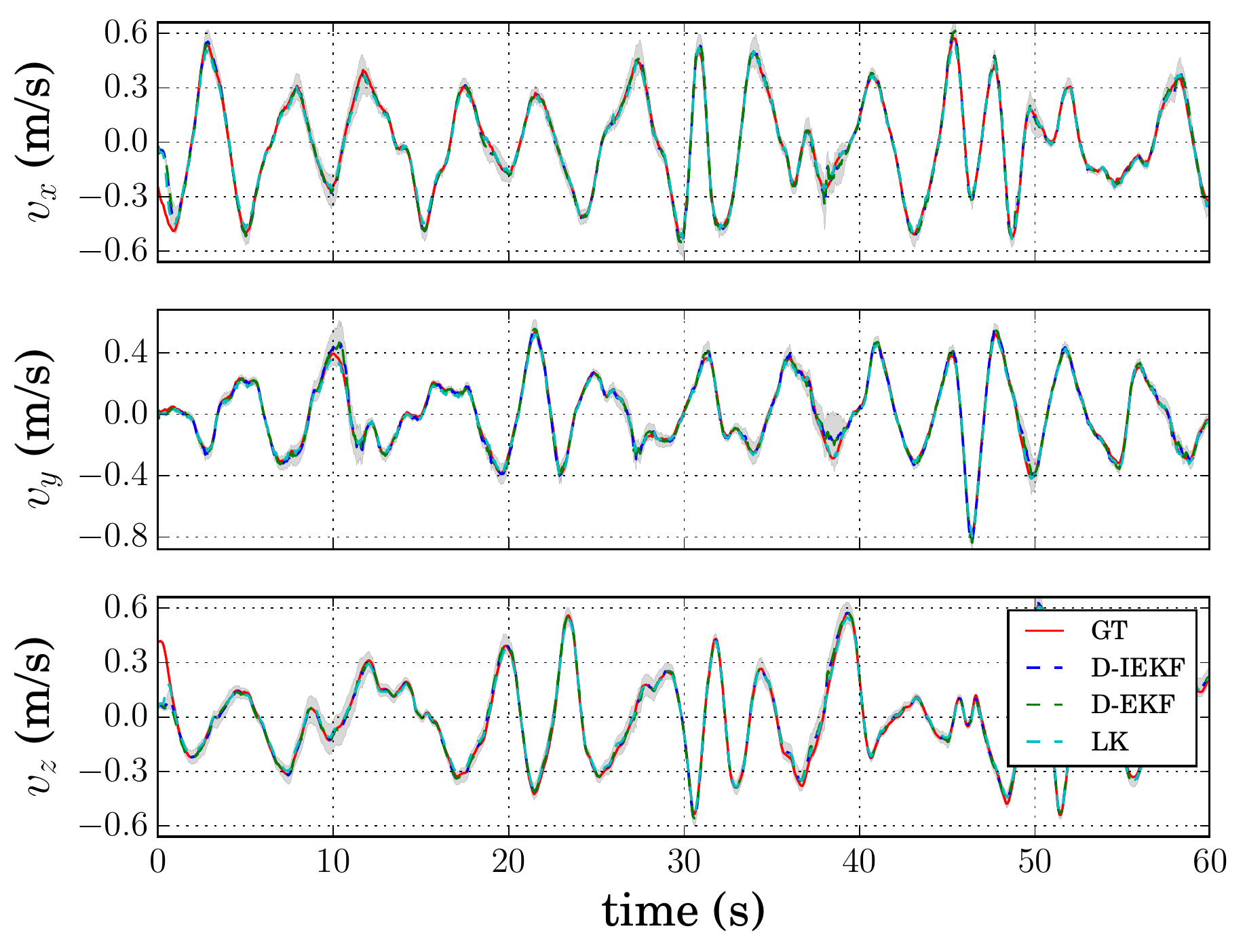}\\
    		(b)\\
    		\includegraphics[scale=0.4]{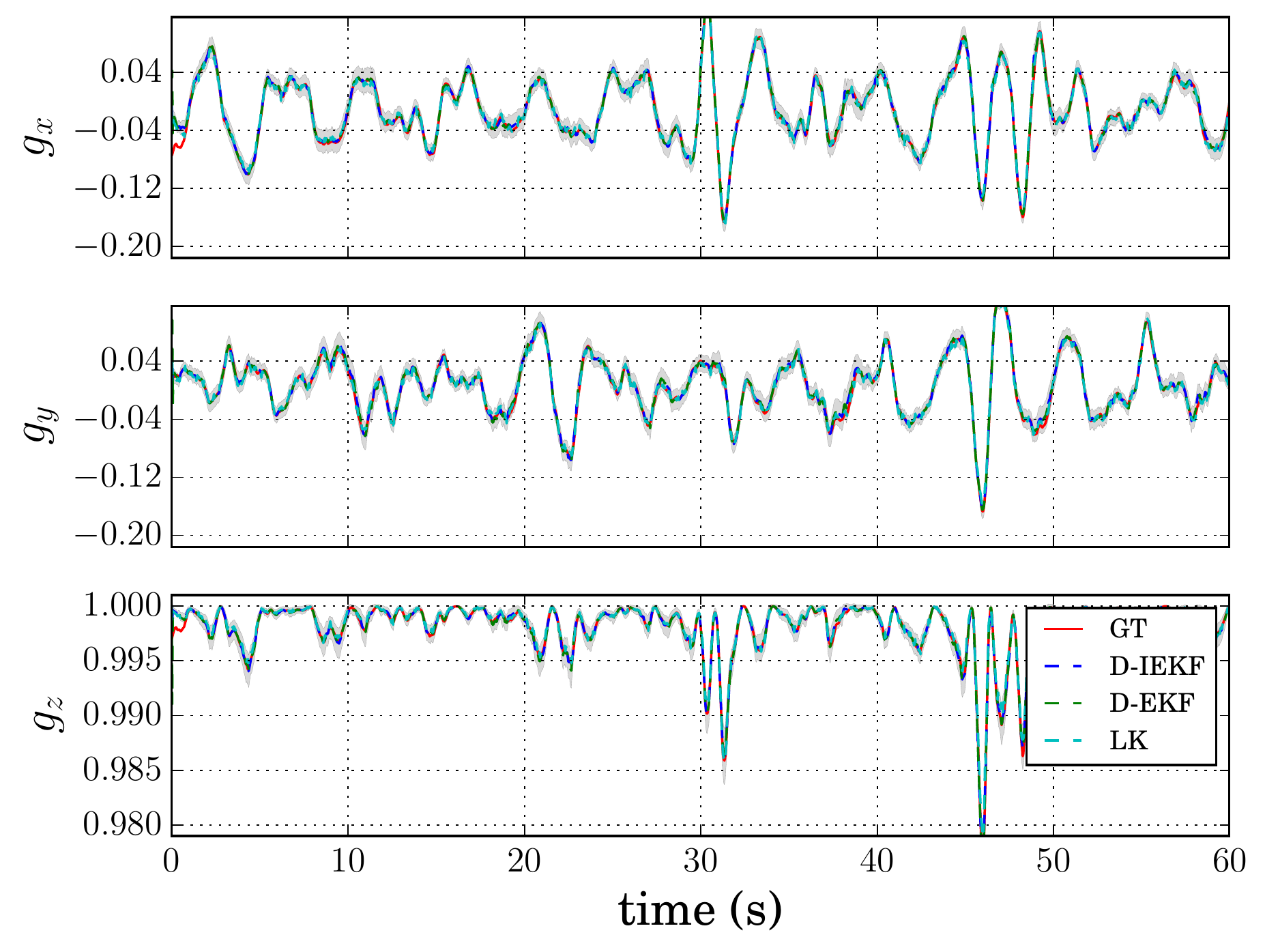} \\
    		(c)\\
    		\includegraphics[scale=0.4]{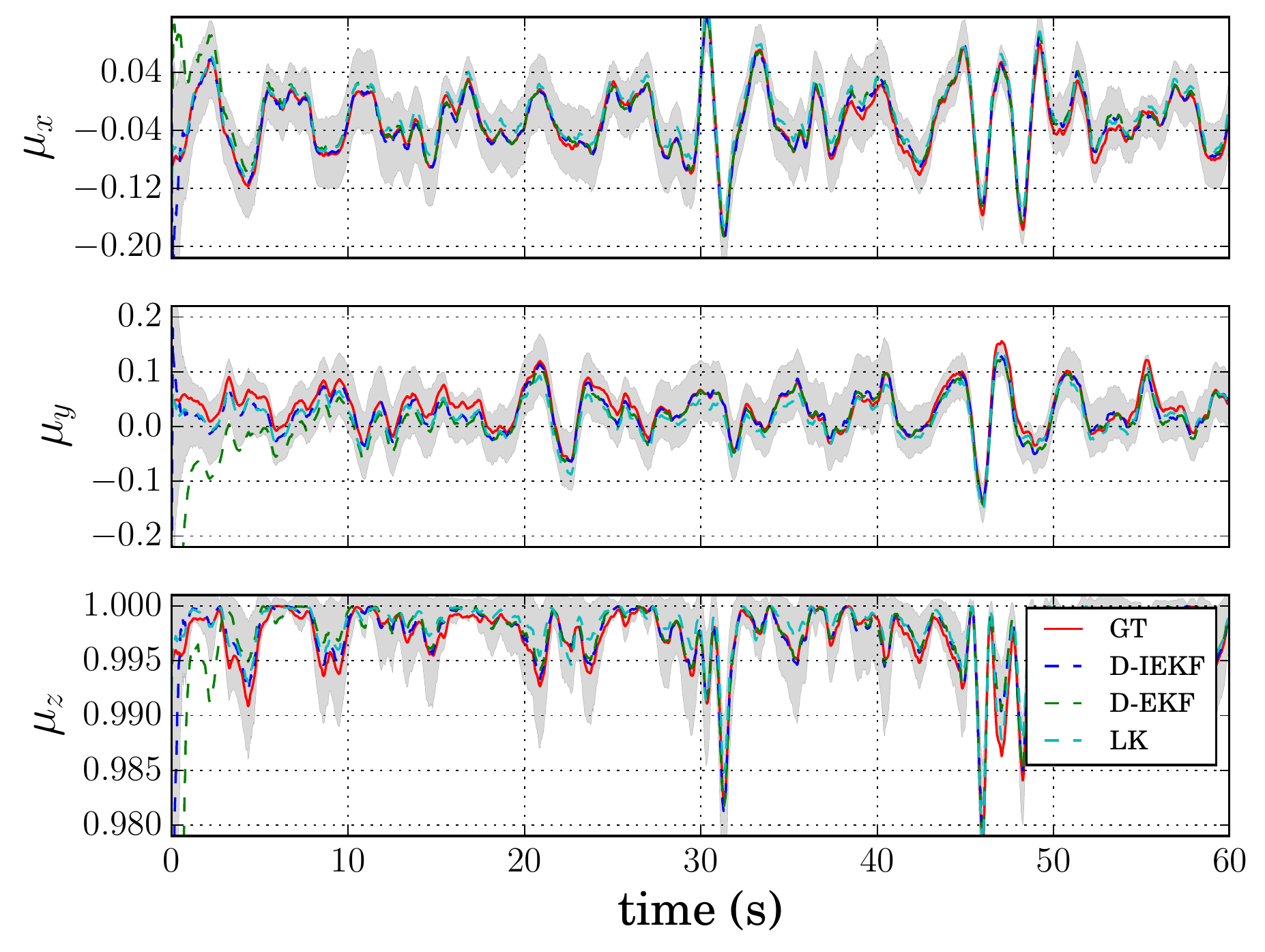}\\
    		(d)
    \end{tabular}
    \caption{Comparison of the estimates from the proposed and benchmak methods using dataset \textcircled{\footnotesize{1}}. The estimates of (a) altitude, (b) vVelocity (c) gravity vector and (d) normal vector from the four approaches are plotted against the ground-truth values (GT). Gray shaded areas indicate 2-$\sigma$ bounds of the D-IEKF estimates.}
	\label{fig:compare_lk_direct}
\end{figure}

The use of the photometric difference between the consecutive frames as IEKF innovation term is one key feature of the proposed method. The featureless approach has the potential to be more robust as it is not susceptible to feature tracking errors. To verify this, the proposed direct method is compared with the traditional LK method. In the implementation of the estimation algorithms, for the proposed direct method (D-IEKF), original images were downsampled to 90 $\times$ 58px. For the LK method, the pipeline is similar to \cite{qin2017vins}. That is, 50 Harris corners \cite{Shi2002Good} were extracted from original images ($752\times480$px). These corners were then tracked by the pyramidal LK method over consecutive images with $20\times 20$ patch size and three image levels. The innovation term in the proposed method is replaced with the difference between the predicted and measured feature coordinates. Our preliminary findings reveal that for the LK method, the update step is vulnerable to tracking outliers, resulting in occasional divergence. To resolve this, the objective function in Eq. \eqref{cost_fun} is robustified with the Huber loss \cite{sibley2009adaptive} for the LK method. In addition to D-IEKF and LK, we also performed the estimation using photometric feedback with a standard EKF (this is equivalent to setting the maximum iteration step of the IEKF to one), notated as D-EKF. For all cases, the weights between the prediction and the image measurements were tuned to obtain the best results for all methods and retained the same for all experiments. 

For assessment, all estimation errors are computed after the estimates converge to the ground-truth, this corresponds to three seconds after the first image update is performed. Table \ref{tab:results_lk_direct} shows the RMSEs of the estimated altitude, linear velocity with respect to the ground-truth, and the average time consumption per frame from all three implementations. It can be seen that both direct methods produce lower RMSEs in the linear velocity and altitude than LK while they are approximately five times faster. It can be concluded that the direct methods outperform the LK method in terms of accuracy and efficiency. This is because, in many circumstances, the quality of the LK estimates suffers from the unreliability caused by incorrect feature correspondences in certain frames. The LK feature association fails to handle repetitive textures such as the CKB pattern, despite the use of Huber loss function, and subsequently corrupts the estimation. This issue could be further ameliorated with an additional outlier rejection strategy such as an application of the epipolar constraint between image correspondences \cite{qin2017vins}. On the other hand, in direct methods, the consistent homography projective constraint is imposed (Eq. \eqref{eq:dense_mea_2}), yielding an inherent outlier rejection. As a result, by exploiting the single plane assumption, the proposed strategy is more robust than the LK method. 

The results from D-EKF exhibit marginally larger RMSEs in the distance and linear velocity compared to that of D-IEKF, consistent with the outcomes in \cite{rovio}. As anticipated, the deprivation of the iterated update defers the convergence of the estimates towards the ground-truth only at the beginning in most cases. The comparable average time cost between these two methods similarly indicates that multiple iterative steps occur almost exclusively at the inception phase of all sequences. After convergence, only one iterative step is required to meet the termination criterion. IEKF essentially accelerates the convergence at a slight increase in computational cost. In addition to the favorable speed-up of the convergence, the result from flight \textcircled{\footnotesize{7}}, in which the robot maneuvered at relatively high speed, highlights the exceptional robustness of D-IEKF. The high flight speed renders the LK method to be extremely inaccurate and causes the D-EKF method to diverge since a single update iteration was not sufficient for the estimation to reduce the initial photometric error between consecutive images and the prediction. The failure to properly find the maximum a posteriori probability estimate (Eq. \eqref{cost_fun}) leads to an accumulation of errors and the divergence of the estimates. This demonstrates that multiple iteration steps enhance the robustness compared to a single iterative step.

Fig. \ref{fig:compare_lk_direct} depicts the estimation results from flight \textcircled{\footnotesize{1}} in detail. The estimated altitudes from all three approaches converge close to the ground truth at around $t=3$ s (Fig. \ref{fig:compare_lk_direct}(a)). The results from D-IEKF and D-EKF are nearly identical after the convergence. Fig. \ref{fig:compare_lk_direct}(a)-(c) reveals that the estimates of the altitude, velocity and gravity direction from all methods are only slightly different whereas the estimates of the plane's normal vector from D-IEKF and D-EKF in Fig. \ref{fig:compare_lk_direct}(d) display a noticeable distinction at the beginning. This corroborates the claim that the iterative update expedites the convergence. Furthermore, it can be observed that the uncertainties of the D-IEKF estimates, and likewise estimation errors, are more pronounced at the extreme points of the velocity plots (Fig. \ref{fig:compare_lk_direct}(b)). These points coincide with the periods where the camera's acceleration approaches zero. This is consistent with the fact that the scale ambiguity of the monocular vision cannot be resolved in the absence of acceleration.

\subsubsection{Comparison of the Proposed Direct Method with the State-of-the-art VINS}{\label{ssec:compare_vins}}
We further compare the D-IEKF method against two state-of-the-art VINS: VINS-Mono \cite{qin2017vins} and ROVIO \cite{rovio} using their published  C++ codes. Unlike the proposed estimator for reactive navigation, VINS-Mono and ROVIO are map-based VINS suitable for autonomous navigation. Since these map-based VINS do not assume the camera to be pointing toward a single flat terrain, the estimate of the distance to a flat terrain or flight altitude is not immediately available for comparison. Despite substantial differences in assumptions and computational complexity, both regimes provide the estimated flight velocity that can be directly compared. This serves as a surrogate measure for comparison of distance estimation owing to the tightly coupled dynamics of velocity and distance. 

VINS-Mono is a variant of a visual-inertial SLAM system rather than a front-end. It features an accurate joint optimization of visual inertial information, loop closure, and map merging and reuse \cite{qin2017vins}. For comparison, the estimates of flight velocity from the sliding window estimator were logged out. In contrast to  VINS-Mono, ROVIO is characterized as a robust and fast visual-inertial front-end. It leverages an IEKF framework by tightly integrating patch-based photometric feedback as its Kalman innovation term. For comparison with the D-IEKF estimates, we used the default ROVIO parameter configuration, which has been well-tuned to achieve a balanced trade-off between accuracy and efficiency. The number of tracked features per frame is set to 25 and the patch size to $6\times 6$. The second and third levels are employed for tracking the multiple level features. 

\begin{figure}
    \centering
    \begin{tabular}{c}
        \includegraphics[scale = 0.45]{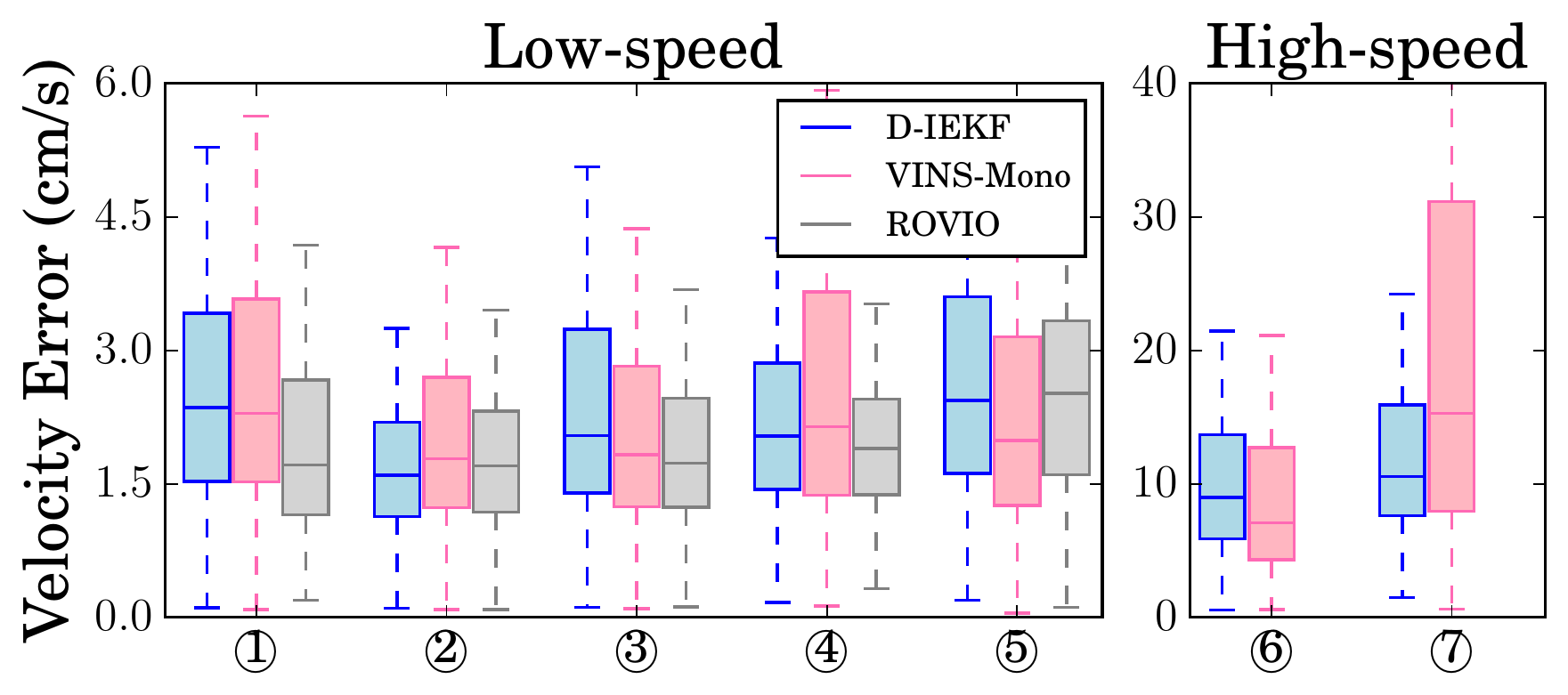} 
    \end{tabular}
    \caption{Boxplots comparing D-IEKF, VINS-Mono, and ROVIO in terms of velocity errors. The left sub-figure shows the results from five low-speed flights and the right sub-figure corresponds to two high-speed flights. The increased flight speed deteriorates the estimation performance. ROVIO estimates failed to converge in the case of high-speed flights.}
    \label{fig:comparison_mynt_rovio_vins}\vspace{-1mm}
\end{figure}

Seven datasets collected from the previous section were used to obtain the velocity estimates from both ROVIO and VINS-Mono as outlined. Fig. \ref{fig:comparison_mynt_rovio_vins} presents the velocity estimation errors using boxplots, depicting the medians and quartiles of the errors. According to the plot, the results demonstrate no overall significant distinction between the three methods in the case of five low-speed maneuvers. Nevertheless, for the two high-speed flights, ROVIO failed the initialization and subsequent tracking as a result. VINS-Mono, on the other hand, has a robust and complex initialization procedure that provides relatively accurate initial estimates. For D-IEKF, the adoption of entire images and iterated updates improve the robustness to deal with the high-speed flights. From the obtained results, it can be concluded that D-IEKF has comparable performance to that of two state-of-the-art VINS when it comes to the flight velocity estimation.
 
In terms of the efficiency, the time consumption per frame averaged from all sequences from all three methods are D-IEKF-1.42 ms, VINS-Mono-41.52 ms, and ROVIO-23.5 ms. Note that for VINS-Mono three threads operate in parallel and only the time cost of the sliding optimization is counted. While the proposed estimator is approximately 15-30 times faster than VINS-Mono and ROVIO, the exceptional computational efficiency is compromised by the lack of mapping and the requirement of a single observed flat surface. 

\subsection{Flights over Tilted Planes}{\label{ssec:incline_plane}}
Different from \cite{Grabe2015Nonlinear,hua2018attitude,chirarattananon2018direct}, our strategy to separately estimate the gravity direction and the plane's normal allows the proposed method to relax the assumption that camera observes horizontal ground. In other words, it is applicable to flights above an inclined plane. To verify this, additional flight experiments were carried out over surfaces covered by the VEG texture with the angles of inclination up to {30\degree} using identical flight configurations and estimation parameters to the experiments on horizontal ground. 

\begin{figure}
    \centering
    \begin{tabular}{c}
        \includegraphics[scale = 0.45]{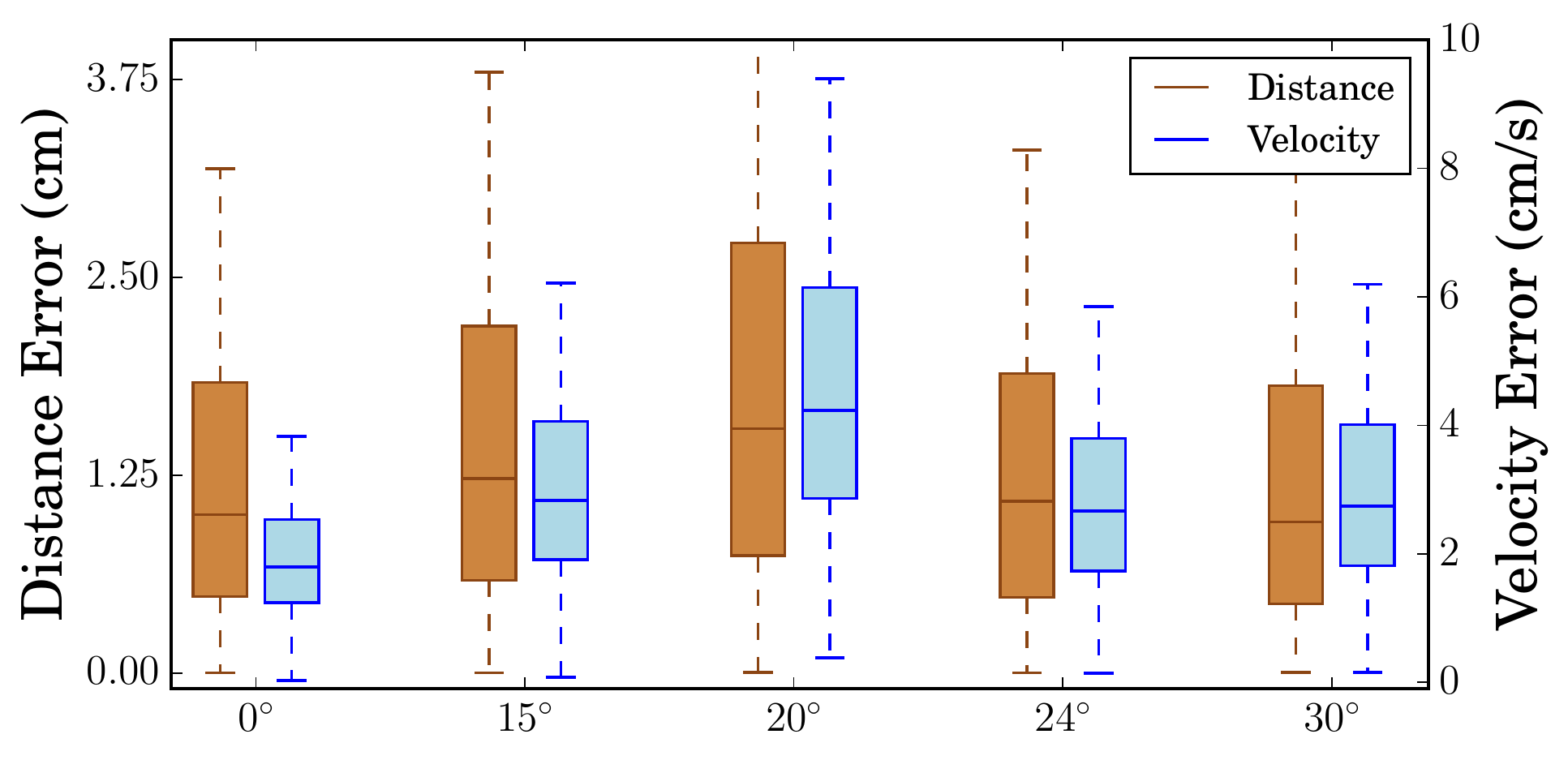} 
    \end{tabular}
    \caption{Distance and velocity errors of the D-IEKF estimates from flights over planes with different angles of inclination. }
    \label{fig:comparison_mynt_angle_rmse}\vspace{-1mm}
\end{figure}
\begin{figure}
    \centering
    \begin{tabular}{c}
        \includegraphics[scale = 0.45]{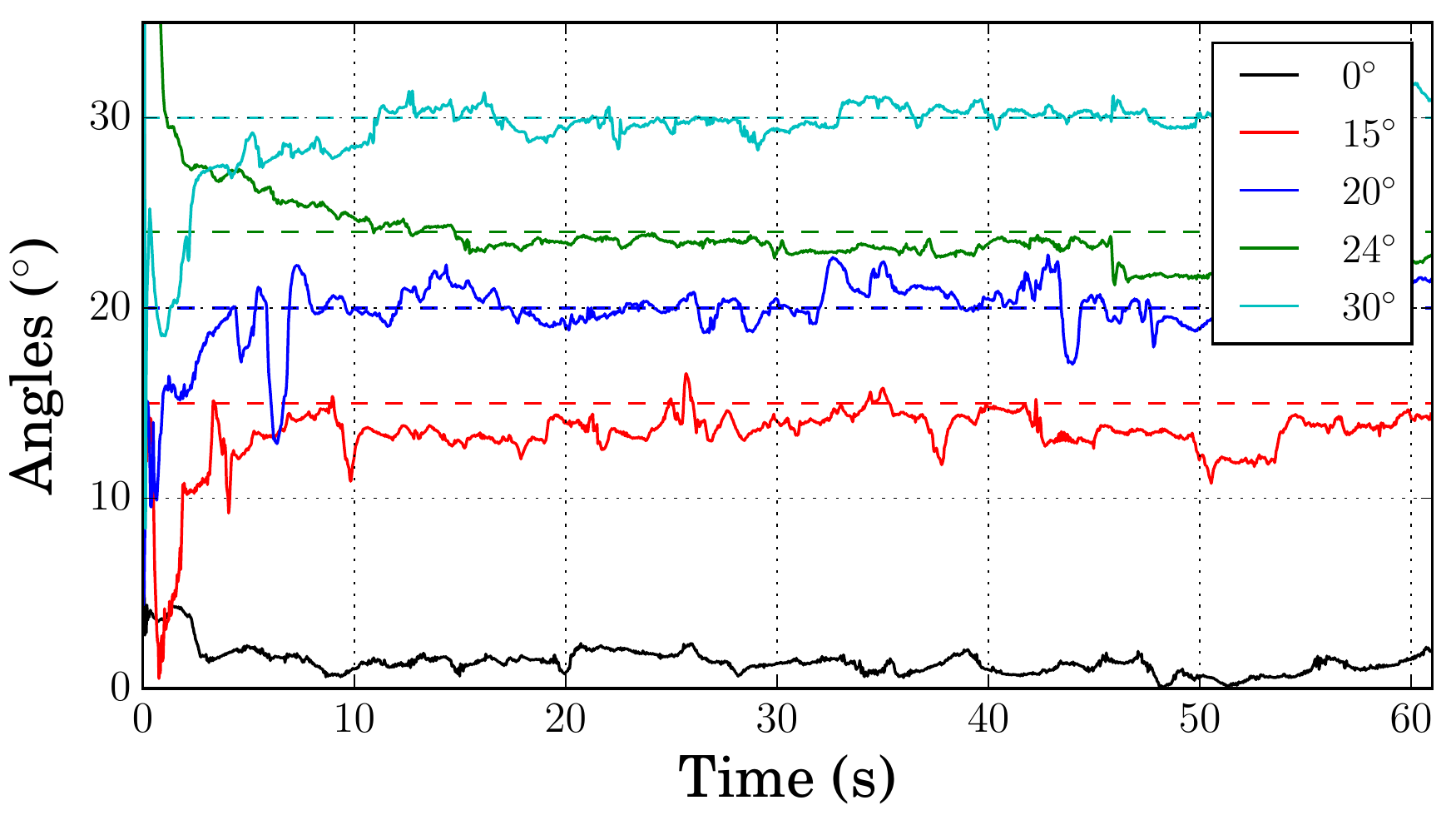} 
    \end{tabular}
    \caption{The angle between the estimated normal vector and gravity vector on different incline planes. The dash lines are the ground-truth angles of the corresponding plane obtained from the motion capture system.}
    \label{fig:comparison_mynt_angle}\vspace{-2mm}

\end{figure}

Fig. \ref{fig:comparison_mynt_angle_rmse} shows the errors of the estimated distance to the inclined planes and the translational velocity. The data belonging to flight \textcircled{\footnotesize{4}} in Table \ref{tab:results_lk_direct} is employed as the flight above a {0\degree}-inclined plane. The plot shows that the accuracy of the estimates is not visibly affected by the plane's inclination. Only marginal variation is seen across different angles. A closer inspection is given in Fig. \ref{fig:comparison_mynt_angle}. The plot demonstrates the angle between the estimated normal vector and the vertical for different cases. The estimated angles evidently oscillate within the vicinity of the true planes' inclination angles. The result from the 20{\degree} case exhibits the largest deviation among the four experiments. That is related to the deteriorated accuracy of the velocity and distance estimates shown in Fig. \ref{fig:comparison_mynt_angle_rmse}, consistent with the relationship predicted by Eq. \eqref{dalpha} and \eqref{d_vartheta}.

\section{Conclusion}
In this paper, we have proposed a computationally efficient framework to estimate the inverse altitude, velocity and the surface's orientation for MAVs from a monocular vision and IMU measurements. The key contribution of our framework lies in the direct use of photometric feedback as the Kalman innovation term. This renders a robust, efficient and inherent data association in a single step. Extensive flight experiments were conducted to demonstrate the effectiveness of our approach. The results prove that the direct use of entire images for feedback offers better accuracy, robustness, and efficiency than the existing feature-based (LK) method. The iterated update scheme improves the estimation with a minimal increase in computation power. Further analysis comparing the proposed method against two state-of-the-art VINS reveals that the accuracy of the velocity estimates calculated from the proposed method is comparable with the two benchmark VINS. It should be highlighted that the exclusion of mapping (and therefore, comprehensive odometry information) and the single plane assumption in the proposed estimator permits it to be $\approx$15-30 times faster than the two VINS. Finally, additional flights were performed to showcase the estimator's ability to determine the plane's normal vector. The results suggest that the achieved estimation performance is not adversely affected when the robot flies over non-horizontal surfaces. Overall, this work offers an attractive lightweight navigation solution for aerial robots with limited computational power.

Possible future directions include the extension of the framework to be applicable with the observation of multiple planes by dealing with the planar area segmentation and ego-motion estimation in one step.\vspace{-1mm}







\bibliography{mybibfile}
\bibliographystyle{IEEEtran}

\end{document}